\documentclass[pdflatex,sn-mathphys-num]{sn-jnl}

\usepackage{graphicx}%
\usepackage{multirow}%
\usepackage{amsmath,amssymb,amsfonts}%
\usepackage{amsthm}%
\usepackage{mathrsfs}%
\usepackage[title]{appendix}%
\usepackage{xcolor}%
\usepackage{textcomp}%
\usepackage{manyfoot}%
\usepackage{booktabs}%
\usepackage{algorithm}%
\usepackage{algorithmicx}%
\usepackage{algpseudocode}%
\usepackage{listings}%
\usepackage{comment}

\usepackage{url}
\usepackage{booktabs}
\usepackage[most]{tcolorbox}

\usepackage{booktabs}
\usepackage{tabularx}
\usepackage{array}
\usepackage{amssymb}
\newtcolorbox{insightbox}[1]{
    colback=orange!10,      
    colframe=orange!70, 
    colbacktitle=orange!70, 
    title={#1},             
    fonttitle=\sffamily\bfseries\large,
    fontupper=\sffamily,    
    arc=15pt,               
    outer arc=15pt,
    left=15pt,              
    right=15pt,
    top=10pt,
    bottom=10pt,
    boxrule=1.5pt,          
    titlerule=0pt,          
    toptitle=5pt,
    bottomtitle=2pt,
    enhanced,               
}

\theoremstyle{thmstyleone}%
\theoremstyle{thmstyletwo}%

\theoremstyle{thmstylethree}%

\begin{document}

\title[Article Title]{VetClaw: An Edge-Cloud Multimodal Agentic System for Veterinary Disease Screening}


\author*[1]{Syed Mhamudul Hasan}\email{syedmhamudul.hasan@siu.edu}

\author[2]{Anas AlSobeh}\email{anas.alsobeh@uvu.edu}

\author[3]{Hussein Zangoti}\email{hmzangoti@jazanu.edu.sa}

\author[1]{Abdur R. Shahid}\email{shahid@cs.siu.edu}


\affil[1]{\orgdiv{School of Computing}, \orgname{Southern Illinois University}, \orgaddress{\city{Carbondale}, \postcode{62901}, \state{IL}, \country{USA}}}

\affil[2]{\orgdiv{Information Systems and Technology}, \orgname{Utah Valley University}, \orgaddress{\street{Orem}, \city{Orem}, \postcode{84058}, \state{UT}, \country{USA}}}

\affil[3]{\orgdiv{College of Engineering and Computer Science}, \orgname{Jazan University}, \orgaddress{ \country{Saudi Arabia}}}


\abstract{We present VetClaw, an edge–cloud multimodal agentic system for early veterinary disease screening. VetClaw uses a camera module as an edge sensing device and sends captured images, together with optional symptom descriptions, to a server-hosted vision-language model for zero-shot disease classification. The system separates agent interaction from workflow orchestration: OpenClaw provides scheduling, tool access, user interaction, and notification services on the edge device, while LangGraph manages the stateful screening workflow, including input validation, image transmission, model invocation, safety checks, conditional routing, failure handling, and structured logging. This design moves beyond static image classification by enabling the system to collect visual evidence, invoke external models, apply deterministic safety rules, and generate diagnostic-support alerts. Results show that image-only VLM prediction remains limited, whereas symptom-guided and multimodal inputs improve zero-shot classification performance. Thus, VetClaw transforms a static prediction model into a coordinated, safety-aware system that can use tools, manage workflows, handle failures, and escalate uncertain cases.}




\maketitle

\section{Introduction}






Animal biosecurity comprises preventive measures designed to reduce the transmission of infectious diseases among animals. For instance, skin disease is one of the main clinical signs observed in livestock when biosecurity fails and a highly contagious germ spreads through the animal colony~\cite{yadav2023introduction, genemo2023detecting, girmaw2025livestock}. Artificial intelligence has a wide range of applications in veterinary medicine, including automated disease prediction~\cite{pereira2023artificial, bashizadeh2024overview, vardhan2025explainable, shandilya2025ai}. However, veterinary diagnosis presents distinct challenges because animals cannot verbally describe symptoms, and disease presentations can vary across species, anatomy, physiology, behaviour, and baseline health patterns. Generative AI may help address this challenge by integrating multiple sources of observable and contextual evidence that can support early animal disease screening through multimodal reasoning and safety-aware response generation~\cite{eckhardt2025livestock}. Instead of relying on a single data type, generative AI can integrate several sources of evidence. This is similar to how veterinarians combine physical examination, imaging, lab results, and observed behavior in real practice to improve veterinary diagnosis by combining different data types such as medical images, clinical notes, lab reports, behavioral videos, and audio recordings to provide more accurate and holistic diagnostic support for animal healthcare~\cite{hennessey2022artificial, gomes2025review}. Agentic AI extends this capability by shifting from static prediction towards adaptive systems that can reason, use tools, coordinate workflows, and support personalised decision-making~\cite{banerjie2025agentic}.


{\color{blue}




}

To demonstrate the application of Agentic AI, we implement an agentic veterinary monitoring pipeline named VetClaw, using an existing veterinary disease dataset containing animal images and associated symptom information to measure zero-shot disease classification performance. The proposed system is designed for real-world edge deployment, where a Raspberry Pi camera module can capture animal images and transmit them, together with user-provided symptom text, to a server-hosted VLM for automated disease screening.



\section{Background}

\subsection{OpenClaw}

OpenClaw is an open-source autonomous-agent framework that provides an interface between language models, external tools, local scripts, scheduled tasks, and communication channels~\cite{steinberger_openclaw_2025}. 


\subsection{LangGraph}

LangGraph is a graph-based framework for implementing stateful language-model and agent workflows. It represents a workflow as a set of nodes connected through deterministic or conditional transitions where each node performs a defined operation, such as validating an input, calling an external model, checking a safety condition, or recording an execution result~\cite{langgraph2024}. 




\subsection{Multimodal AI}
Multimodal AI refers to systems that can process and reason across multiple types of data such as text, images, audio, and video rather than being limited to a single modality. For instance, a VLM is a specific type of multimodal AI that combines vision for image understanding and language for text generation and reasoning~\cite{soenksen2022integrated}.
\subsection{Related Work}


AI has strong potential for the early detection and management of animal diseases~\cite{feighelstein2024ai, michelena2025review, fuentes2022livestock}. Multimodal AI can extend this potential by integrating visual, textual, physiological, and behavioural signals, making it a promising foundation for veterinary diagnostic support~\cite{bhardwaj2024machine, gomes2025review, attri2025saam, paulauskaite2026ai}. Agentic AI can further strengthen such systems by coordinating tools, maintaining workflow state, handling failures, and supporting adaptive decision processes~\cite{xu2026comprehensive}. Prior work has explored several relevant directions. VetTag~\cite{zhang2019vettag} is a large-scale NLP system for automatically assigning veterinary diagnosis codes from free-text clinical notes. In a similar way, Lee et al.~\cite{lee2025automated} proposed a multi-agent system for clinical problem detection in human healthcare, and Sun et al.~\cite{sun2024conversational} investigated LLM-based diagnosis in human healthcare using external planners and simulated patient dialogues. Although their work did not address veterinary applications. Hennessey et al.~\cite{hennessey2022artificial} reviewed the broader role of AI in veterinary medicine, including diagnostic imaging and disease prediction. Related work has also examined cloud-connected sensing systems for livestock monitoring. For example, Darvesh et al.~\cite{darvesh2023iot} developed an IoT and machine learning platform that uses real-time sensor measurements, including skin temperature, heart rate, and motion, to monitor cattle health, predict potential medical conditions, and notify producers through a mobile application. However, their system does not incorporate an agentic orchestration layer for multimodal reasoning.

{\color{red}

}

\subsection{Contributions}
To fill this gap, this paper makes three contributions. 

\begin{enumerate}

    \item We introduce VetClaw, an OpenClaw-based edge-cloud agentic workflow for veterinary disease screening using images captured by a Raspberry Pi and server-hosted VLM. 


    \item We evaluate two VLMs under image-only, text-only, and text-image zero-shot settings across two public veterinary disease datasets.

    \item We analyze the role of symptom descriptions and multimodal fusion, showing that text-guided VLM reasoning improves performance over image-only prediction while introducing potential leakage and safety risks that require controlled prompting and rule-based review. 
\end{enumerate}

\begin{figure*}[t!]
    \centering
    \includegraphics[width=\textwidth]{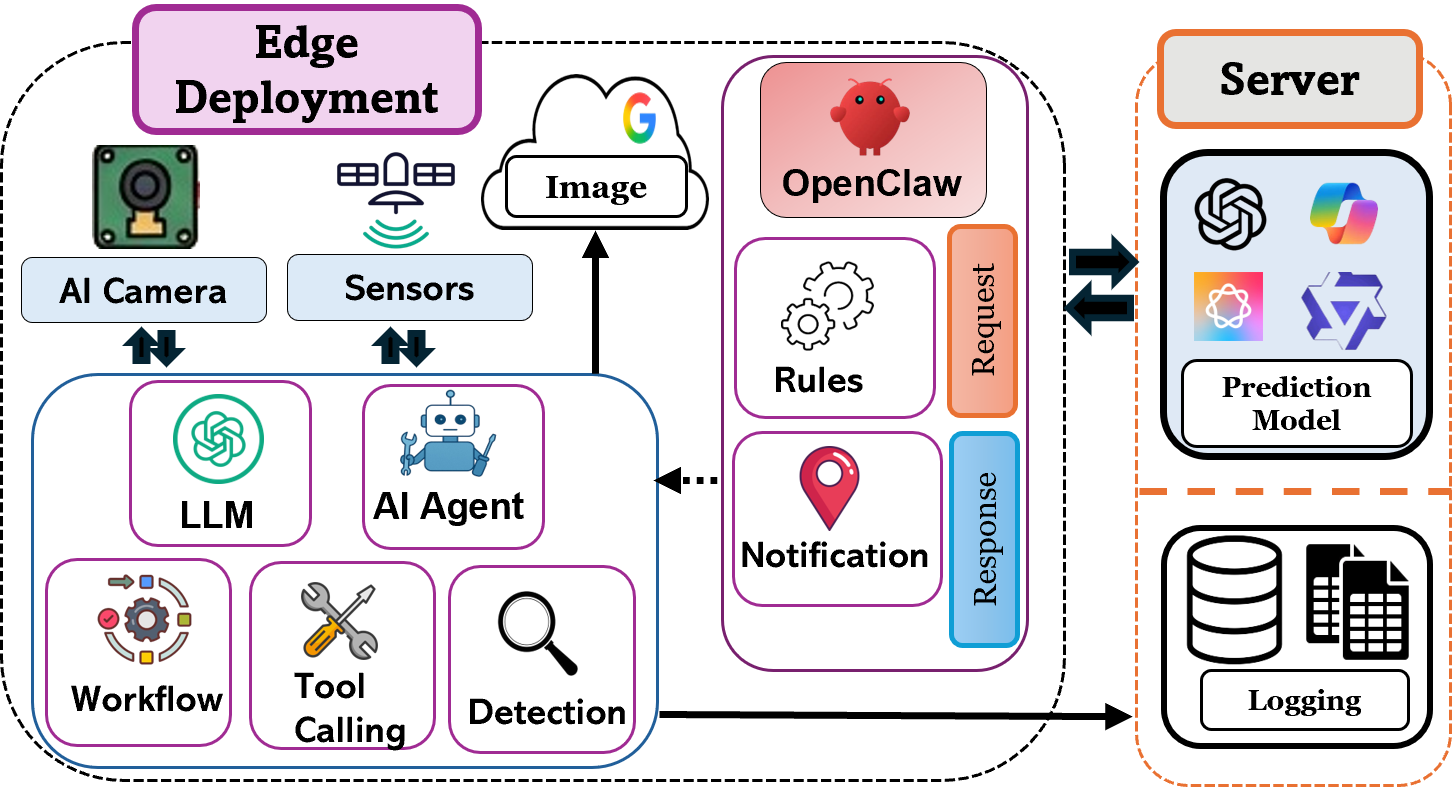}

    \caption{Edge-cloud Architecture of the VetClaw Veterinary Assistant Deployed on a Raspberry Pi for Early Skin Disease Prediction.}
    \label{fig:vetclaw}
\end{figure*}

\section{VetClaw Architecture}

(\textbf{System Component.}) VetClaw consists of an edge deployment layer and a server-side inference layer, as illustrated in Figure~\ref{fig:vetclaw}. The edge layer contains a Raspberry Pi, Camera Module 3, OpenClaw, a lightweight local language model, and a LangGraph workflow. OpenClaw acts as the edge-facing agent. It schedules image capture, invokes local tools, accepts optional symptom information, starts the LangGraph workflow, and delivers screening notifications. A lightweight local language model may support instruction interpretation and user interaction, but it does not perform the final disease classification.
LangGraph operates inside the edge application as the workflow orchestration layer. After OpenClaw starts a screening session, LangGraph maintains the case state and controls the sequence of operations. Its nodes validate the captured image, preprocess the input, upload or encode the image, construct the API request, invoke the server-hosted VLM, validate the returned response, apply safety rules, determine whether escalation is required, and store structured logs.
The server-side layer contains a FastAPI endpoint, a server-hosted VLM, and persistent logging and storage services. FastAPI validates the incoming request and forwards the image and, optionally, the symptom text to the VLM. The VLM returns a structured prediction containing a candidate disease category and supporting information. The result is then returned to the LangGraph workflow for deterministic safety review. The system, therefore, assigns distinct responsibilities to each component. OpenClaw manages interaction, scheduling, tools, and notification. LangGraph controls workflow state, transitions, retries, safety rules, and escalation. FastAPI manages server communication and request validation. The VLM performs multimodal disease classification. Because large VLMs cannot be efficiently deployed on a Raspberry Pi, VetClaw follows an edge–cloud architecture where the Raspberry Pi performs lightweight sensing, workflow initiation, validation, communication, and notification, while computationally expensive VLM inference is performed on a remote server. (\textbf{Agentic Workflow}) VetClaw uses a layered agentic architecture in which OpenClaw and LangGraph perform complementary functions. OpenClaw combined with LangGraph operates as the external agent and tool-access layer, managing task scheduling, camera invocation, symptom collection, workflow initiation, and result delivery. LangGraph operates as the internal stateful workflow engine, controlling input validation, branch selection, image transfer, API invocation, response parsing, safety checking, retries, escalation, and logging. Each operation is represented as a workflow node in LangGraph, while conditional edges determine the next action based on the current case state.

The server-hosted VLM functions as an inference tool within the LangGraph workflow rather than as the workflow controller. It predicts a disease category but does not determine whether the result should be delivered, suppressed, retried, or escalated. This separation improves traceability, reproducibility, and system control. The Raspberry Pi handles sensing, scheduling, local tool execution, workflow initiation, API communication, and notification, while the remote server performs computationally expensive VLM inference.

The workflow begins when the OpenClaw scheduler triggers a screening task for LangGraph which calls a Python-based camera tool to capture an animal image and saves it locally with a timestamp. OpenClaw also invokes the LangGraph workflow and provides the image path and any available symptom text as the initial workflow state. LangGraph validates the image, determines whether symptom information is available, and routes the case to either an image-only or text–image inference branch. After preprocessing, the workflow uploads or encodes the image and sends the image reference and symptom text to the FastAPI endpoint for response. After receiving the VLM response, LangGraph validates the response format and applies deterministic safety rules. Cases involving invalid outputs, low confidence, conflicting evidence, urgent symptoms, or API errors are routed to retry, recapture, follow-up, or human-review nodes. Cases that pass the safety checks are converted into early screening alerts rather than final clinical diagnoses. Finally, LangGraph records a structured execution log and returns the reviewed result to OpenClaw, which presents the alert or sends a notification through the configured communication channel. 

\section{Dataset and Tools}

\quad \textbf{Dataset.} For evaluation, we use two public image datasets: Pet Disease Images~\cite{adive_pet_disease_images} and the Dogs Skin Disease Dataset~\cite{motiani_dogs_skin_disease_dataset}. The Pet Disease Images dataset contains 1,736 images distributed across 22 disease classes. On the other hand, the Dogs Skin disease contains 443 dog images across four classess. We use all the images from both datasets for testing for zero-shot evaluation. 

\textbf{Agentic AI.} VetClaw uses OpenClaw as the edge-facing agent deployed on the Raspberry Pi. OpenClaw schedules screening tasks, invokes local tools, initializes the LangGraph workflow, and communicates the reviewed result to the user. LangGraph executes the stateful screening workflow, including preprocessing, input-mode selection, image transmission, server invocation, rule-based safety checking, and escalation. 

\textbf{Edge Device.} We use a CanaKit Raspberry Pi with 8 GB of RAM, an ARM processor, and 128 GB of storage connected to a Raspberry Pi Camera Module 3. The Raspberry Pi runs Python 3.12, and a Python-based camera script is used to capture animal images, while the sensor generates optional textual data. 
The image and optional sensor text data are then provided to the disease-prediction models that are hosted on a remote server and accessed through a FastAPI endpoint. 

\textbf{Server.} We deploy two VLMs: Qwen3-VL-32B~\cite{bai2025qwen3} and InternVL3-38B~\cite{zhu2025internvl3} for vision-based zero-shot classification. The models were hosted on a server equipped with an Intel Xeon processor, an NVIDIA A100 GPU with 80 GB of memory, and 240 GB of system RAM. Our VLMs support text-only prompting, which does not require an image during prompting. Afterward, we evaluate these VLMs in three input settings: image-only, text-only, and text-image. The text-only setting is included as an ablation baseline to measure the contribution of language priors and disease-description cues without visual input, and it helps distinguish whether the VLM's performance is driven primarily by visual evidence, textual reasoning, or multimodal fusion. To avoid label leakage, we removed disease names, folder names, file paths, and class-identifying metadata from all prompts. The text-only condition used only symptom-style descriptions without the ground-truth class label where the model was prompted to select one label from a fixed candidate list. 

\textbf{API Communication.} The VLM is hosted on a server using Ollama, while a Python FastAPI layer exposes the model through a REST API. The sensor of Raspberry Pi collects predefined textual data, while the Raspberry Pi camera module captures animal images. LangGraph then sends the image and optional symptom text to the FastAPI endpoint. The server then forwards the request to the Ollama-hosted VLM, receives the model response, and returns the predicted disease class and safety-guided diagnostic-support output to the LangGraph workflow.


\begin{table}[t]
\centering
\small
\setlength{\tabcolsep}{3.2pt}
\renewcommand{\arraystretch}{1.08}
\caption{Zero-shot classification performance of Qwen3-VL-32B and InternVL3-38B under image-only, text-only, and text–image settings across two veterinary disease datasets. In practice, we should use only fine-tuned VLMs for better results. However, we provide the results of two state-of-the-art VLMs to show that text and images together can yield better performance.}
\label{tab:dataset}
\begin{tabularx}{\linewidth}{llXcccc}
\toprule
\textbf{Dataset} & \textbf{VLM} & \textbf{Input} & \textbf{Acc. (\%)} & \textbf{Prec.} & \textbf{Rec.} & \textbf{F1} \\
\midrule

\multirow{6}{*}{\shortstack[l]{Dogs Skin\\Disease}}
& \multirow{3}{*}{Qwen3-VL-32B}
& Image only   & 33.26  & 0.51 & 0.31 & 0.31 \\
& & Text only    & 69.00  & 0.60 & 0.75 & 0.64 \\
& & Text + image & 72.17  & 0.85 & 0.78 & 0.69 \\
\cmidrule(lr){2-7}
& \multirow{3}{*}{InternVL3-38B}
& Image only   & 31.90  & 0.41 & 0.35 & 0.31 \\
& & Text only    & 100.00 & 1.00 & 1.00 & 1.00 \\
& & Text + image & 94.34  & 0.94 & 0.95 & 0.94 \\

\midrule

\multirow{6}{*}{\shortstack[l]{Pet Disease\\Images}}
& \multirow{3}{*}{Qwen3-VL-32B}
& Image only   & 48.41 & 0.64 & 0.48 & 0.47 \\
& & Text only    & 86.90 & 0.79 & 0.86 & 0.82 \\
& & Text + image & 84.02 & 0.90 & 0.83 & 0.80 \\
\cmidrule(lr){2-7}
& \multirow{3}{*}{InternVL3-38B}
& Image only   & 52.22 & 0.62 & 0.52 & 0.49 \\
& & Text only    & 86.04 & 0.79 & 0.86 & 0.81 \\
& & Text + image & 88.11 & 0.93 & 0.88 & 0.86 \\

\bottomrule
\end{tabularx}
\end{table}

\section{System Implementation and Performance Evaluation}

\subsection{Workflow Initialization}

The VetClaw workflow is initialized by the OpenClaw scheduler at predefined intervals on the Raspberry Pi. In this implementation, OpenClaw acts as the edge-facing agent and initializes the LangGraph workflow on the Raspberry Pi and triggers a Python-based camera script to capture an animal image using the Raspberry Pi Camera Module 3. LangGraph then validates the captured image before passing it to the preprocessing stage. If symptom text is available from the sensor, the workflow combines the image with the symptom description to form a multimodal input for disease screening. After input preparation, OpenClaw sends the image and optional symptom text to the server-side API via LangGraph. The server forwards the request to the hosted VLM for disease prediction and returns the model response to the Raspberry Pi. Once the response is received, LangGraph applies rule-based safety checks to identify uncertain, high-risk, or potentially urgent cases. These rules prevent the system from presenting the VLM output as a final diagnosis and instead frame the result as an early screening alert. For traceability, each workflow execution in LangGraph generates a server-side log containing the timestamp, image reference, request status, model response, predicted disease category, and rule-checking outcome. Captured images are stored in Google Drive for future review and analysis, while workflow logs are maintained on the server. If the rule engine detects a suspicious or urgent condition, the system generates a notification that includes the screening message and the corresponding captured image to review the visual evidence and seek professional veterinary support.

\subsection{Image Processing}
Image processing begins when OpenClaw invokes LangGraph workflow, which initiate a Python camera tool on the Raspberry Pi. The Camera Module 3 captures an animal image and stores it locally. A LangGraph node then validates that the image exists, can be opened, and satisfies basic quality requirements. Additional checks of LangGraph include image resolution, file size, blur level, brightness, and supported file type and images that fail validation are routed to a recapture node rather than being sent to the VLM. After validation, the image is prepared for server-side inference. The workflow may upload the image to Google Drive and send a secure image reference, or it may encode the image in a request-compatible format. In the current prototype, the image is uploaded to Google Drive, and its reference is added to the LangGraph state. LangGraph does not need to transfer a raw image object between every workflow node. Instead, it maintains the local path, cloud reference, or encoded representation as part of the structured workflow state. The selected image representation is then sent to the FastAPI endpoint together with the optional symptom description. This design separates image capture, workflow control, and model inference. LangGraph manages image validation and transmission, and the remote server performs computationally expensive VLM inference.

\begin{table*}[t]
\centering
\caption{Sample execution log for the VetClaw execution cycle including capture, upload, request, and response stages. Note: Image size and latency may vary depending on image resolution, network conditions, cloud upload time, and server-side processing load. Latency was measured on a server equipped with an Intel Xeon processor, an NVIDIA A100 GPU with 80 GB, and 240 GB of system RAM. Latency may vary across hardware configurations and network conditions.  }
\begin{tabular}{lrr}
\toprule
\textbf{Stage} & \textbf{Field} & \textbf{Value} \\
\midrule
\multirow{4}{*}{Capture}
  & Device    & Camera Module 3 \\
  & Timestamp & 2026-06-03 \\
  & Status    & Success \\
  & Size      & 296 KB \\
  & Time & 2820.47 ms \\
\midrule
\multirow{4}{*}{Upload}
  & Destination & Google Drive \\
  & File ID     & 20260603\_125355.jpg \\
  & Status      & Success \\
  & Time   & 1977.86 ms \\
\midrule
\multirow{4}{*}{Response}
  & Model        & Qwen3-VL-32B \\
  & Prediction   & Dental Disease in Cat \\
  & Time*     & 925.3 ms \\
\bottomrule
\end{tabular}
\label{tab:request_response}

\end{table*}

\subsection{VLM Performance}

Table~\ref{tab:dataset} summarizes the zero-shot classification performance of the two VLMs. Overall, image-only performance is weak, while text-guided and multimodal settings achieve stronger results. For example, the performance on the Dogs Skin Disease Dataset improves from a macro-F1 score of 0.31 in the image-only setting to 0.69 macro-F1 in the text-image mode, and the model achieves a macro-F1 of 0.82 on the Pet Disease Images dataset. Across both datasets, image-only performance was limited, while the text-only and text–image settings achieved stronger macro-F1 scores, indicating that disease descriptions and contextual cues play an important role in veterinary AI prediction. In the Dogs Skin Disease Dataset, the text-image setting achieved the best overall performance, with 72.17\% accuracy and 0.69 macro-F1. In the Pet Disease Images dataset, the text-only setting achieved the highest accuracy of 86.90\%, while the text-image setting produced strong macro precision of 0.90, showing the potential of VLM-based reasoning for disease screening. These findings suggest that multimodal veterinary diagnosis should not rely only on images but should combine visual evidence with symptom descriptions, veterinary rules, and contextual knowledge through the agentic AI.

\subsection{Request \& Response}
After image preprocessing, LangGraph constructs a structured JSON request for server-side inference that contains the image reference, optional sensor observations, timestamp, etc. Afterward, the FastAPI endpoint validates the required fields and prepares the multimodal input for the server-hosted VLM. The VLM returns a structured response containing the predicted disease category, supporting evidence, uncertainty information, and a diagnostic-support explanation. The returned response is not sent directly to the user. It is first passed back to the LangGraph workflow. A response-validation node checks whether the output follows the expected schema and whether the predicted category belongs to the permitted candidate set. A safety node then evaluates uncertainty, urgent symptoms, inconsistent image–text evidence, unsupported cases, and failure conditions. LangGraph produces a reviewed workflow result containing the model prediction, safety status, escalation decision, and notification message. It also writes the request, response, rule outcomes, errors, and timestamps to the structured log. The final reviewed result is returned to OpenClaw, which presents the alert or sends a notification containing the screening message and associated image. To evaluate real-time feasibility, VetClaw logs latency and resource metrics for each request-response cycle shown in Table~\ref{tab:request_response}. The edge device records image capture time, local save time, upload time, request payload size, etc. The server records image retrieval time, VLM inference latency, safety-check latency, and total server-side processing time. The final response also includes API round-trip latency and total end-to-end latency from image capture to screening alert. These measurements allow us to compare image transfer strategies, such as URL-based upload and Base64 encoding, and to identify whether latency is dominated by edge capture, network transfer, Google Drive, or VLM inference.

\section{Limitations}
This study has several limitations that should be considered when interpreting the results. First, the experiments were conducted using two relatively small veterinary image datasets. Although the selected datasets provide useful initial evidence for evaluating the proposed VetClaw workflow, their size and diversity are limited. 
Second, this study relies on zero-shot prompting rather than supervised fine-tuning. Zero-shot evaluation is useful for testing the general reasoning ability of VLM models without task-specific training. However, it may not capture fine-grained veterinary disease patterns as reliably as a model trained directly on labeled veterinary data. The model’s predictions can also be sensitive to prompt wording, disease label descriptions, and the amount of contextual information provided. As a result, zero-shot performance should be interpreted as an early feasibility result, not as final clinical-grade diagnostic performance. Third, the results depend strongly on the selected VLMs (e.g. Qwen3-VL-32B and InternVL3-38B) for zero-shot classification across image-only, text-only, and text-image settings. However, different VLMs may produce different outcomes depending on their training data, visual reasoning ability, medical knowledge, and instruction-following quality. A stronger or more specialized veterinary/medical VLM may yield better classification results, while a smaller or weaker VLM may reduce accuracy and reliability. 
Fourth, VetClaw can capture only images from a fixed direction due to its camera module. More advanced cameras, such as AI camera modules, can make it more suitable for capturing images of moving animals in real-world deployment. Finally, there is a broader lack of high-quality veterinary multimodal data. The current study uses image datasets, but real veterinary diagnosis often requires multiple evidence sources, including clinical notes, lab results, behavioral observations, audio, video, and physical examination findings. 


\section{Conclusion}
This study presented VetClaw for early animal disease screening using edge device, a sensor, a camera module and server-hosted VLM. The system separates edge interaction from workflow orchestration: OpenClaw provides scheduling, tool access, user interaction, and notification, while LangGraph manages the stateful screening workflow, including preprocessing, model invocation, response validation, safety checking, conditional escalation, failure handling, and structured logging. The experimental results suggest that symptom-guided and multimodal inputs can improve zero-shot veterinary disease classification compared with image-only prediction. Future work will include larger-scale workflow testing, comparison against non-agentic pipelines, repeated latency measurements, safety-rule evaluation, failure-recovery experiments, and validation by veterinary professionals before real-world clinical use.

\section*{Funding}

No funding was received for this study.

\section*{Ethical Considerations }

This study does not involve any sensitive data and uses only publicly available veterinary image datasets that do not involve any private veterinary records, owner information, or personally identifiable data. 

\section*{Acknowledgment}

This paper presents a work-in-progress project. The authors remain responsible for the technical content, experimental design, analysis, and final manuscript decisions.

\bibliography{sn-bibliography}

\end{document}